# Situations and Computation: An Overview of Recent Research*

Erkan Tın and Varol Akman
Department of Computer Engineering and Information Science
Faculty of Engineering, Bilkent University
Bilkent, 06533 Ankara, Turkey
{tin,akman}@bilkent.edu.tr

**Abstract**

Serious thinking about the computational aspects of situation theory is just starting. There have been some recent proposals in this direction (viz. PROSIT and ASTL), with varying degrees of divergence from the ontology of the theory. We believe that a programming environment incorporating bona fide situation-theoretic constructs is needed and describe our very recent BABY-SIT implementation. A detailed critical account of PROSIT and ASTL is also offered in order to compare our system with these pioneering and influential frameworks.

## 1  Introduction

*Situation theory* has been devised to develop a unified mathematical theory of meaning and information content and to clarify and resolve various long-standing problems in the study of language, information, logic, philosophy, and the mind. The original theory was due to Jon Barwise and John Perry [6]. The theory has matured over the last decade [12] and various versions of it have been applied to a number of linguistic issues, resulting in what is commonly known as *situation semantics* [2, 3, 5, 11, 15, 16, 18, 25]. Situation semantics aims at the construction of a mathematically rigorous theory of meaning, and the application of such a theory to natural language.

The mathematical foundations of the theory are based on intuitions basically coming from set theory and logic [1, 3, 11, 12]. One of the distinguishing characteristics of situation theory vis-à-vis another influential semantic theory in the logical tradition [13] is that information content is context-dependent.

*This work has been supported in part by a NATO Science for Stability Grant (TU-LANGUAGE) awarded to Bilkent University.

While not much work has been done to construct a computational framework based on situation theory, there have been some attempts to investigate this [8, 9, 16, 23]. These have incorporated only some of the original features of the theory; the remaining features were omitted for the sake of achieving particular goals. This has caused conceptual and philosophical divergence from the ontology of the theory. Recent studies [26, 27, 28, 29] have tried to avoid this pitfall by simply sticking to the essentials of the theory and adopting the ontological features which were first put forward by Barwise and Perry [6] and then refined by Devlin [12].

In this paper, we review the existing approaches towards a computational account of situation theory; this may serve as a guideline for researchers who aim at constructing programming systems permitting the use of situation-theoretic constructs. We briefly review situation theory and situation semantics in Sections 2 and 3. In Section 4, we briefly explain why situations should be used in natural language processing, and in knowledge representation for semantic interpretation and reasoning. The existing computational accounts based on situation theory are examined in Section 5. Finally, we present our concluding remarks in Section 6.

## 2  Situation Theory

Individuals, properties, relations, spatio-temporal locations, and situations are basic constructs of situation theory. The world is viewed as a collection of objects, sets of objects, properties, and relations. *Individuals* are conceived as invariants; having properties and standing in relations they persist in time and space. All individuals, including spatio-temporal locations, have *properties*.

*Infons* [12] are discrete items of information and *situations* are first-class objects which describe parts of the real world. Infons are denoted as $\ll R, a_1, \ldots, a_n, i \gg$ where $R$ is an $n$-place relation, $a_1, \ldots, a_n$ are objects appropriate for the respective argument places of $R$, and $i$ is the polarity (0 or 1). Situations are intensional objects. For this reason, abstract situations are proposed to be their counterparts amenable to mathematical manipulation. An abstract situation is defined as a set-theoretic construct. Given a real situation $s$, the set $\{\sigma \mid s \models \sigma\}$, where $\sigma$ is an infon, is the corresponding abstract situation. Here, $s$ is said to *support* $\sigma$ (denoted as $s \models \sigma$) just in case $\sigma$ is true of $s$. Otherwise, $s \not\models \sigma$. In case of abstract situations, the *supports* relation reduces to set-inclusion.

Suppose Alice was eating ice cream yesterday at home and she is also eating ice cream now at home. Both of these situations share the same constituent sequence $\ll$eats, Alice, ice cream$\gg$. These two events, occurring at the same location but at different times, have the same *situation type s*. Situation types are partial functions from relations and objects to 0 and 1. The situation type $s$ in our example assigns 1 to the constituent sequence $\ll$eats, Alice, ice cream$\gg$.

Situation types can be more general. For example, a situation type in which someone is eating something at home 'contains' the situation in which Alice is eating ice cream at home. Suppose Alice is not present in the room where this paper is being written. Then, "Alice is eating ice cream" is not part of our situation $s$ and hence gets no truth value in $s$. Thus, situation theory allows *partiality* [16].

Situations in which a sequence is assigned both truth values are *incoherent*. For instance, a situation $s$ is incoherent if $\ll$has, Alice, A$\heartsuit$, 0$\gg \in s$ and $\ll$has, Alice, A$\heartsuit$, 1$\gg \in s$. This is a situation in which Alice has the A$\heartsuit$ and she does not have the A$\heartsuit$ in a card game. There cannot be a real situation $s$ validating this. Nevertheless, the constituent sequence $\ll$has, Alice, A$\heartsuit\gg$ may be assigned these truth values for spatio-temporally distinct situation types (say, $s$ and $s'$).

Allowing partiality has the advantage of distinguishing between logically equivalent statements. For example, the statements "Bob is angry" and "Bob is angry and Bob is shouting or Bob is not shouting" are logically equivalent in the classical sense. In situation semantics, these two sentences will not have the same interpretation. A situation $s$ describing the situation in which Bob is only angry will not contain any sequence about Bob's shouting, i.e., $s$ will be 'silent' on Bob's shouting. However, another situation $s'$ obtained as the union of two situations (Bob is angry and Bob is shouting; Bob is angry and Bob is not shouting) will contain a sequence about Bob's shouting.

A *scheme of individuation*, a way of carving the world into uniformities, is an essential aspect of situation theory. The notions of individual, relation, spatial and temporal location depend upon this schema of individuation. In other words, the basic constituents of the theory are determined by the agent's schema of individuation. Formal representation of these uniformities yields what is known as *types*.

Situation theory provides a collection of basic types that can be used for individuating or discriminating uniformities of the real world. There are nine basic types: temporal location (TIM), spatial location (LOC), individual (IND), $n$-place relation (REL$^n$), situation (SIT), infon (INF), type (TYP), parameter (PAR), and polarity (POL).

If $R$ is an $n$-place relation and $a_1, ..., a_m$ ($m \leq n$) are objects appropriate for the argument places $i_1, ..., i_m$ of $R$, and if the filling of these argument places is sufficient to satisfy the minimality conditions for $R$, then for $i \in \{0, 1\}$, $\ll R, a_1, ..., a_m, i \gg$ is a well-defined *infon*. *Minimality conditions* for a particular relation are the collection of conditions that determine which particular groups of argument roles need to be filled in order to produce an infon. If $m < n$, the infon is said to be *unsaturated*; if $m = n$ it is *saturated*.

*Abstraction* can be captured in a primitive level by allowing parameters in infons. Parameters are generalizations over classes of non-parametric objects (e.g., individuals, spatial locations). Parameters of a parametric object can be associated with objects which, if they were to replace the parameters, would yield one of the objects in the class that parametric object abstracts over. The parametric objects actually define types of objects in that class. Hence, allowing parameters in infons results in *parametric infons*. For example, ≪sees, $\dot{x}$, Alice, 1≫ and ≪sees, $\dot{x}$, $\dot{y}$, 1≫ are parametric infons where $\dot{x}$ and $\dot{y}$ stand for individuals. These infons are said to be parametric on the first, and the first and second argument roles of the relation *sees*, respectively.

*Anchoring* (binding) parameters of an infon to real objects yields *parameter-free infons*. For example, in ≪goes, $\dot{x}$, Chicago, 1≫ if an anchoring function $f$ anchors $\dot{x}$ to the individual John, we obtain the parameter-free infon ≪goes, John, Chicago, 1≫.

Parameters can be restricted so that they represent finer uniformities. Given a parameter $\dot{x}$ and a set of infons $I$, $\dot{x}\hat{} I$ restricts the class of objects that can be anchored to $\dot{x}$ only to the ones for which $I$ hold in the 'world' situation. This process is known as *parameter restriction*.

Complex object types can be defined over some intial situation. Given a situation $s$, a parameter $\dot{x}$, and a set of infons (involving $\dot{x}$) $I$, one can define $[\dot{x} \mid s \models I]$ to denote the type (class) of all objects for which the conditions imposed by $I$ hold in $s$. This process of obtaining a type from a parameter $\dot{x}$, a situation $s$, and a set $I$ of infons, is referred to as *type-abstraction*. $\dot{x}$ is called the *abstraction parameter* while $s$ is called the *grounding situation*.

A situation $s'$ is *part-of* another situation $s$ (denoted by $s' \triangleleft s$) just in case $(\forall \sigma)[s' \models \sigma \Rightarrow s \models \sigma]$. The *part-of* relation between situations is anti-symmetric, reflexive, and transitive, and consequently provides a partial-ordering of the situations.

In situation theory, information flow is made possible by a network of abstract 'links' between high-order uniformities, viz. *situation types*. These links are called *constraints*. Barwise and Perry identify three forms of constraints [6]. *Necessary constraints* are those by which one can define or name things, e.g., "Every dog is a mammal." *Nomic constraints* are patterns that are usually called natural laws, e.g., "Blocks fall if not supported." *Conventional constraints* are those arising out of explicit or implicit conventions that hold within a community of living beings, e.g., "The first day of the month is the pay day." They are neither nomic nor necessary, i.e., they can be violated. All types of constraints can be *conditional* and *unconditional*. Conditional constraints can be applied to situations that fulfill some condition while unconditional constraints can be applied to all situations.

## 3 Situation Semantics

According to situation semantics, meanings of expressions reside in systematic relations between different types of situations. They can be identified with relations on *discourse situations* $d$, *(speaker) connections* $c$, the utterance $\varphi$ itself, and the described situation $e$. Some public facts about $\varphi$ (such as its speaker and time of utterance) are determined by the discourse situations. The ties of the mental states of the speaker and the hearer with the world constitute $c$. A discourse situation involves the expression uttered, its speaker, the spatio-temporal location of the utterance, and the addressee(s). Each of these defines a linguistic role (the role of the speaker, the role of the addressee, and so on).

The utterance of an expression $\varphi$ 'constrains' the world in a certain way, depending on how the *roles* for discourse situations, connections, and described situations are occupied. For example, "I am crying" describes a three-place relation 〚I am crying〛 on the utterance situation (the discourse situation and the connections) $u$ and the described situation $e$. This defines a meaning relation:

$$d, c \llbracket \text{I am crying} \rrbracket e.$$

Given a discourse situation $d$, connections $c$, and a course of events $e$, this relation holds just in case there is a location $l_d$ and a speaker $a_d$ such that $a_d$ is speaking at $l_d$, and in $e$, $a_d$ is crying at $l_d$.

In interpreting the utterance of an expression $\varphi$ in a context $u$ $(d, c)$, there is a flow of information, partly from the linguistic form encoded in $\varphi$ and partly from contextual factors provided by the utterance situation $u$. These are combined to form a set of constraints (not uniquely determined) on the described situation $e$: given $u$ and an utterance of $\varphi$ in $u$, there will be several situations $e$ that satisfy the constraints imposed. While the meaning of an utterance of $\varphi$ and hence its interpretation are influenced by other factors such as stress, modality, and intonation [16], the situation in which $\varphi$ is uttered and the situation $e$ described by this utterance seem to play the most influential roles. For this reason, the meaning of an utterance is essentially taken to be a relation defined over $\varphi$, $u$, and $e$. This approach towards identifying linguistic meaning is essentially what Barwise and Perry call the *Relation Theory of Meaning* [6].

Situation semantics makes simple assumptions about the way natural language works. Primary among them is the assumption that language is used to convey information about the world (the so-called *external significance* of language). Even when two sentences have the same interpretation, i.e., they describe the same situation, they can carry different information.

Classical approaches to semantics underestimate the role played by *context-dependence*; they ignore pragmatic factors such as intentions and circumstances of the individuals involved in the communicative process. But, indexicals, demonstratives, tenses, and other linguistic devices rely heavily on context for their interpretation. Context-dependence is an essential hypothesis of situation semantics. A given sentence can be used over and over again in different situations to say different things (the so-called *efficiency* of language). Its interpretation, i.e., the class of situations described by the sentence, is therefore subordinate on the situation in which the sentence is used. This context-providing situation, *discourse situation*, is the speech situation, including the speaker, the addressee, the time and place of the utterance, and the expression uttered. Since speakers are always in different situations, having different causal connections to the world and different information, the information conveyed by an utterance will be relative to its speaker and hearer (the so-called *perspectival relativity* of language) [6].

Besides discourse situations, the interpretation of an utterance depends on the speaker's connections with objects, properties, times and places, and on the speaker's ability to exploit information about one situation to obtain information about another. Therefore, context supports not only facts about speakers, addressees, etc. but also facts about the relations of discourse participants to other contextually relevant situations such as *resource situations*. Resource situations are contextually available and provide entities for reference and quantification.

According to situation semantics, we use meaningful expressions to convey information not only about the external world but also about our minds (the so-called *mental significance* of language). Situation semantics differs from other approaches in that we do not, in attitude reports, describe our mind directly (by referring to states of mind, ideas, senses, thoughts, etc.) but indirectly (by referring to situations that are external).

With these underlying assumptions and features, situation semantics provides a fundamental framework for a realistic model-theoretic semantics of natural language. The ideas emerging from research in situation semantics have been coalesced with well-developed linguistic theories such as *lexical-functional grammar* and led to rigorous formalisms [16]. On the other hand, situation semantics has been compared to other influential mathematical approaches to the theory of meaning, viz. Montague Grammar [10, 13, 24] and *Discourse Representation Theory* (DRT) [19].

## 4 Why Compute with Situations?

A computational formulation of situation theory may generate interest among artificial intelligence and natural language processing researchers. The theory claims that its model theory is more amenable to a computationally tractable implementation than standard model theory (of predicate calculus) or Montague Grammar. This is due to the fact that situation theory emphasizes partiality whereas standard model theory is clearly holistic.

From a natural language processing point of view, situation theory is interesting and relevant simply because the linguistic account of the theory (viz. situation semantics) handles various linguistic phenomena with a flexibility that surpasses other proposals. It seems that indexicals, demonstratives, referential uses of definite descriptions, pronouns, tense markers, names, etc. all have technical treatments in situation semantics that reach beyond available theoretical apparatuses. For example, the proposed mechanisms, as reported in [18], for dealing with quantification and anaphoric connections in English sentences are all firmly grounded in situation semantics. The insistence of situation semantics on contextual interpretation makes the theory more compatible with speech act theory (and pragmatics in general) than other theories.[1]

# 5 Computational Frameworks

## 5.1 PROSIT

PROSIT (PROgramming in SItuation Theory) is the pioneering work in this direction. PROSIT is a situation-theoretic programming language developed by Nakashima et al. [23]. It has been implemented in Common Lisp.

PROSIT is tailored more for knowledge representation in general than for natural language processing. One can define situations and assert knowledge in particular situations. It is also possible to define relations between situations in the form of constraints. PROSIT's computational power is due to an ability to draw inferences via rules of inference. There is an inference engine similar to a Prolog interpreter. PROSIT offers a treatment of partial objects, such as situations and parameters. It can also deal with self-referential expressions [5].

One can assert facts that a situation will support. For example, if `s1` supports the fact that Bob is a young person, this can be defined in the current situation `s` as:

```
s:  (!= s1 (young Bob)).
```

Note that the syntax is similar to that of Lisp and the fact is in the form of a predicate. The supports relation, `!=`, is situated so that whether a situation supports a fact depends on where the query is made.

In PROSIT, there exists a tree hierarchy, with the situation `top` at the root of the tree. `top` is the global situation and the 'owner' of all the other situations generated. One can traverse the 'situation tree' using the predicates `in` and `out`. Although it is possible to make queries from a situation about any other situation, the result will depend on where the query is made. If a situation `sit2` is defined in the current situation, say `sit1`, then `sit1` is said to be the owner of `sit2`.

The owner relation states that if `(!= sit2 infon)` holds in `sit1`, then `infon` holds in `sit2`, and conversely, if `infon` holds in `sit2` then `(!= sit2 infon)` holds in `sit1`. So, `in` causes the interpreter to go to a specified situation which will be a part of the 'current situation' (the situation in which the predicate is called) and `out` causes the interpreter to go to the owner of the current situation.

---

[1] Kamp's DRT may safely be considered as the only competition in this regard [19]. However, it should be noted that there are currently research efforts towards providing an 'integrated' account of situation semantics and DRT, as witnessed by Barwise and Cooper's recent work [4].

Similar to the owner relation between situations there is the 'subchunk' relation. It is denoted as (c< sit1 sit2), where sit1 is a subchunk of sit2, and conversely, sit2 is a superchunk of sit1. When a situation, sit1, is asserted to be the subchunk of another situation, sit2, it means that sit1 is totally described by sit2. A superchunk is like an owner except that out will always cause the interpreter to go to the owner, not to a superchunk.

PROSIT has two more relations that can be defined between situations. These are the 'subtype' relation and the 'subsituation' relation. When the subtype relation (denoted by (-> sit1 sit2)) is asserted, it causes the current situation to describe that sit1 supports i for every infon i valid in sit2 and that sit1 respects every constraint that is respected by sit2, i.e., sit2 becomes a subtype of sit1. The subsituation relation is denoted as (s< sit1 sit2) and is the same as (-> sit1 sit2) except that only infons, but no constraints, are inherited. Both relations are transitive.

One can define a 'default inheritance' relation between two situations. When a default inheritance relation (denoted by (d< sit1 sit2)) is asserted, sit1 inherits an infon i to sit2 if and only if (no i) cannot be proved to hold in sit2.

The fact that PROSIT permits situations as arguments to infons makes it possible to represent self-referential statements. Consider a card game where there are two players. John has the ace of spades and Mary has the queen of spades. When both players display their cards the following infons will be true:

    (!= sit (has John ace-of-spades))
    (!= sit (has Mary queen-of-spades))
    (!= sit (sees John sit))
    (!= sit (sees Mary sit))

There is no notion of situation type in PROSIT. For this reason, one cannot represent abstractions over situations and specify relations between them without having to create situations and assert facts to them.

It is possible to define a relation as an abstraction over parameters of an infon. A PROSIT expression of the form

    [ $par_1$ ... $par_n$ | $infon$ ]

describes an abstraction and it can be applied to arguments:

    ([ $par_1$ ... $par_n$ | $infon$ ] $arg_1$ ... $arg_n$)

to yield $infon'$ where $infon'$ is the result of replacing each $par_i$ in $infon$ with the corresponding $arg_i$. Therefore, abstraction in PROSIT does not yield an object type or situation type in the situation-theoretic sense.

PROSIT allows definition of a special kind of infon which is called *restricted infon*. An expression of the form

    (^ $infon1$ $infon2$)

defines an infon where *infon2* is the restriction of *infon1*. For example,

    (^ (man P) (human P))

puts a restriction on the parameter `P` of the infon `(man P)` such that `P` must fulfill the relation `human`. Hence, a restriction specifies what relations hold of the parameters of the infon. This approach does not provide a mechanism equivalent to parameter restriction; rather it seems to offer a limited mechanism to specify appropriateness conditions for a given relation and a specific parameter.

PROSIT has a constraint mechanism. Constraints can be specified using either of the three relations $\Rightarrow$, $\Leftarrow$, and $\Leftrightarrow$. Constraints specified using $\Rightarrow$ (respectively, $\Leftarrow$) are forward (respectively, backward) chaining constraints; the ones using $\Leftrightarrow$ are both backward- and forward-chaining constraints. Backward chaining constraints are of the form ($\Leftarrow$ *head fact$_1$ ... fact$_n$*). If all the facts are supported by the situation, then the head fact is supported by the same situation. Forward chaining constraints are of the form ($\Rightarrow$ *fact tail$_1$ ... tail$_n$*). If *fact* is asserted to the situation, then all the tail facts are asserted to the same situation. Backward chaining constraints are activated at query-time while forward-chaining constraints are activated at assertion-time. By default, all the tail facts of an activated forward-chaining constraint are asserted to the situation, which may in turn activate other forward-chaining constraints recursively.

For a constraint to be applicable to a situation, the situation must be declared to 'respect' the constraint. This is done by using the special relation *respect*. For example, to state that every man is human, one would write:

    s:  (resp s1 (<= (human *X) (man *X))).

This states that `s1` respects the stated constraint and is made with respect to `s`. (`*X` denotes a variable.) Since assertions are situated, a situation will or will not respect a constraint depending on where the query is made. If we assert:

    s:  (!= s1 (man Bob)),

then PROSIT will affirmatively answer the query:

    s?  (!= s1 (human Bob)).

Constraints in PROSIT are about local facts within a situation rather than about situation types. That is, the interpretation of constraints does not allow direct specification of constraints between situations, but only between infons within situations.

Parameters, variables, and constants are used for representing entities in PROSIT. Variables, rather than parameters, are used to identify the indeterminates in a constraint. Parameters might be used to refer to unknown objects in a constraint. Variables have a limited scope; they are local to the constraint in which they appear. Parameters, on the other hand, have global scope. Variables match any expression in the language and parameters be can equated to any constant or parameter.

PROSIT has been used to show how problems involving cooperation of multiple agents can be studied, especially by combining reasoning about situations. In [22], Nakashima et al. demonstrate how the *Conway paradox* [3, pp. 201–220] can be solved. The agents involved in this problem use the *common knowledge* accumulated in a shared situation. This situation functions as a communication channel containing all information known to be commonly accessible. One agent's internal model of the other is represented by situations. Individual knowledge situation plus the shared situation help an agent to solve the problem; also cf. [14] for further work on similar *epistemic puzzles*.

## 5.2  ASTL

Black's ASTL (A Situation Theoretic Language) is another programming language based on situation theory [8]. ASTL is aimed at natural language processing. One can define in ASTL constraints and rules of inference over the situations. An interpreter, a basic version of which is implemented in Common Lisp, passes over ASTL definitions and answers queries about the set of constraints and basic situations.

ASTL allows individuals, relations, situations, parameters, and variables. These form the basic terms of the language. Complex terms are in the form of i-terms (to be defined shortly), situation types, and situations. Situations may contain facts which have those situations as arguments. Sentences in ASTL are constructed from terms and can be constraints, grammar rules, or word entries.

The complex term *i-term* is simply an infon $\langle rel, arg_1, \ldots, arg_n, pol \rangle$ where $rel$ is a relation of arity $n$, $arg_i$ is a term, and $pol$ is either 0 or 1. A *situation type* is given in the form $[par \mid cond_1 \ldots cond_n]$ where $cond_i$ has the form $par \models$ *i-term*. If situation S1 supports the fact that Bob is a young person, this can be defined as:

```
S1:   [S | S |= <young,bob,1>].
```

The single colon indicates that S1 supports the situation type on its right-hand side. The supports relation in ASTL is global rather than situated. Consequently, query answering is independent of the situation in which the query is issued.

Constraints are actually backward-chaining constraints. Each constraint is of the form $sit_0 : type_0 \Leftarrow sit_1 : type_1, \ldots, sit_n : type_n$, where $sit_i$ is a situation or a variable, and $type_i$ is a situation type. If each $sit_i$, $1 \leq i \leq n$, supports the corresponding situation type, $type_i$, then $sit_0$ supports $type_0$. For example, the constraint that every man is a human being can be written as follows:

```
*S: [S | S |= <human,*X,1>] <= *S: [S | S |= <man,*X,1>].
```

*S, *X are variables and S is a parameter. An interesting property of ASTL is that constraints are global. Thus, a new situation of the appropriate type need not have a constraint explicitly added to it. Assume that S1, supporting the fact that Bob is a man, is asserted:

    S1:   [S | S $\models$ ⟨man,bob,1⟩].

This together with the constraint above would give:

    S1:   [S | S $\models$ ⟨human,bob,1⟩].

Grammar rules are another form of constraint. An example grammar rule describing the utterance of a sentence consisting of a noun phrase and verb phrase is

    *S: [S | S $\models$ ⟨cat,S,sentence,1⟩] $\rightarrow$
        *NP: [S | S $\models$ ⟨cat,S,nounphrase,1⟩],
        *VP: [S | S $\models$ ⟨cat,S,verbphrase,1⟩]

where cat denotes the category of the construct, and $\rightarrow$ indicates that this is a grammar rule. The rule reads: "When there is a situation *NP of the given type and situation *VP of the given type, there is also a situation *S of the given type."

As in PROSIT, variables in ASTL have scope only within the constraint they appear. They match any expression in the language unless they are declared to be of some specific situation type in the constraint. Hence, it is not possible to declare variables as well as parameters to be of other types such as individuals, relations, etc. Consequently, anchoring of parameters cannot be achieved appropriately in ASTL.

The primary motivation underlying ASTL was to figure out a framework in which semantic theories such as situation semantics [3] and DRT [19] can be described and possibly compared. Such an attempt can be found in [7].

## 5.3 Situation Schemata

Situation schemata have been introduced by Fenstad et al. [16] as a theoretical tool for extracting and displaying information relevant for semantic interpretation from linguistic form. The boundaries of situation schemata are flexible and depending on the underlying theory of grammar, are susceptible to amendment.

A simple sentence $\varphi$ has the situation schema shown in Figure 1(a). Here $r$ can be anchored to a relation, and $a$ and $b$ to objects; $i \in \{0,1\}$ gives the polarity. LOC is a function which anchors the described fact relative to a discourse situation $d, c$. LOC will have the general format in Figure 1(b). IND.$\alpha$ is an indeterminate for a location, $r$ denotes one of the basic structural relations on a relation set $R$, and $loc_0$ is another location indeterminate. The notation $[\ ]_\alpha$ indicates repeated reference to the shared attribute value, IND.$\alpha$. A partial function $g$ anchors the location of SIT.$\varphi$, viz. SIT.$\varphi$.LOC, in the discourse situation $d, c$ if

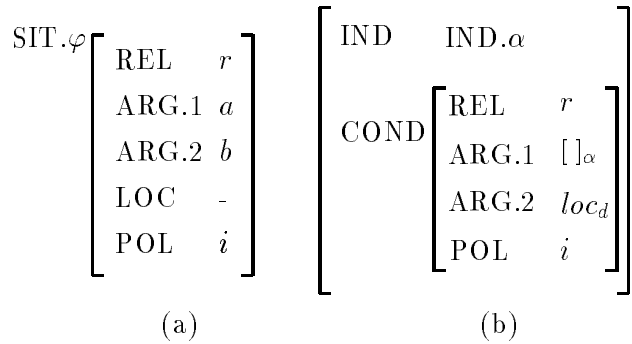

Figure 1: (a) A prototype situation schema, (b) the general format of LOC in (a).

$g(loc_0) = loc_d$ and

$c(r), g(\text{IND}.\alpha), loc_d; 1$

where $loc_d$ is the discourse location and $c(r)$ is the relation on $R$ given by the speaker's connection $c$. The situation schema corresponding to "Alice saw the cat" is given in Figure 2.

Situation schemata can be adapted to various kinds of semantic interpretation. One could give some kind of operational interpretation in a suitable programming language, exploiting logical insights. But in their present form, situation schemata do not go further than being complex attribute-value structures. Situations, locations, individuals, and relations constitute the basic domains of the structure. Constraints are declarative descriptions of the relationships holding between aspects of linguistic form and the semantic representation itself.

Theoretical issues in natural language semantics have been implemented on pilot systems employing situation schemata. The grammar described in [16], for example, has been fully implemented using a lexical-functional grammar system [17] and a fragment including prepositional phrases has been implemented using DPATR.

### 5.4 BABY-SIT

BABY-SIT is a computational medium based on situations, a prototype of which is currently being developed in $\text{KEE}^{TM}$ (Knowledge Engineering Environment) [20]. The implementation language is Common Lisp and the BABY-SIT desktop is based on X Windows running on a SPARCStation (cf. Figure 3). The primary motivation underlying BABY-SIT is to facilitate the development and testing of programs in domains ranging from linguistics to artificial intelligence in a unified framework built upon situation-theoretic constructs [26, 27, 29]. An interactive environment helps one to develop and test his program, observe its behavior

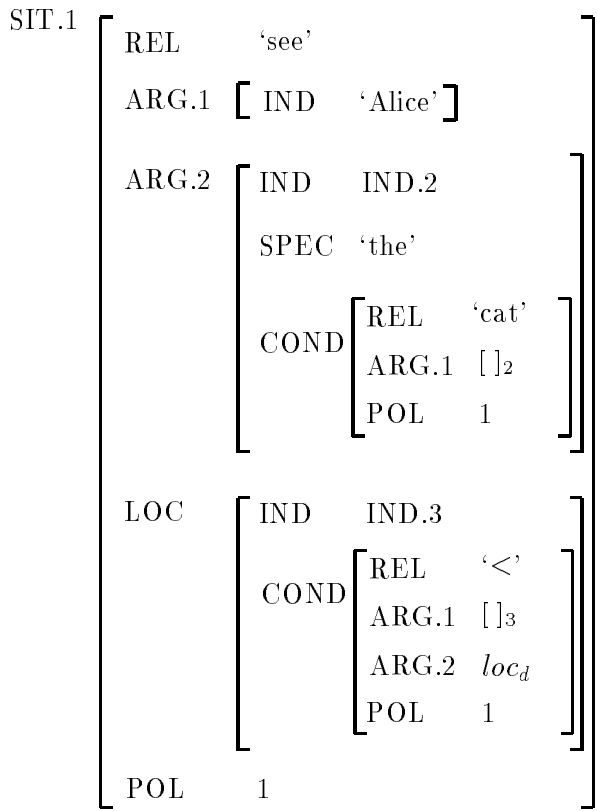

Figure 2: Situation schema for "Alice saw the cat."

vis-à-vis extra (or missing) information, and issue queries [26].

The computational model underlying the current version of BABY-SIT consists of nine primitives: *individuals, times, places, relations, polarities, parameters, infons, situations,* and *types.* Each primitive carries its own internal structure. *Individuals* are unique atomic entities which correspond to real objects in the world. *Times* are individuals of distinguished type, representing temporal locations and, similar to times, *places* are individuals which represent spatial locations. A relation has argument roles which must be occupied by appropriate objects. *Infons* are the discrete items of information of the form $\ll rel, arg_1, \ldots, arg_n, pol \gg$, where $rel$ is a relation, $arg_i$, $1 \leq i \leq n$, is an object of the appropriate type for the $i$th argument role, and $pol$ is the polarity. *Parameters* are 'place holders' for objects in the model. They are used to refer to arbitrary objects of a given type. *Types*, on the other hand, form higher-order uniformities for individuating or discriminating other uniformities in the world.

*Situations* are set-theoretic constructs, e.g., a set of *parametric infons* (comprising relations, parameters, and polarities). A parametric infon is the basic computational unit. By defining a hierarchy between them, situations can be embedded via the special relation *part-of.* In this way, a situation $s$ can have information about another situation $s'$ which is part of $s$. A distinguished situation called *background situation* (denoted by $w$) contains infons which are inherited by all situations, i.e., $w$ is implicitly part of all situation structures in the environment and its infons hold in all situations. However, situations other than $w$ may contain infons that vary from situation to situation. A situation can be either (spatially and/or temporally) *located* or *unlocated*. Time and place for a situation can be declared by *time-of* and *place-of* relations, respectively.

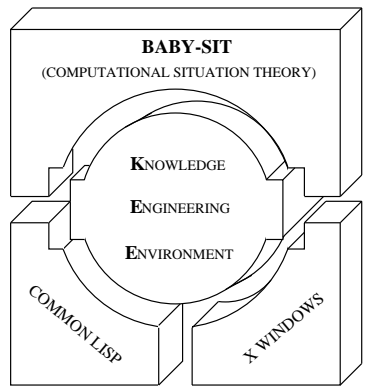

Figure 3: The software structure of BABY-SIT.

Anchoring of parameters is made possible by *anchoring situations* which allow parameters to be anchored to objects of appropriate types—individuals, situations, parameters, etc. But a parameter must be anchored to a unique object in an anchoring situation, i.e., it is anchored once in a given anchoring situation. On the other hand, more than one parameter may be anchored to the same object in an anchoring situation. Anchoring of a parameter can be done via the special relation *anchor*. Restrictions on parameters must be satisfied by $w$.

There are three modes of computation in BABY-SIT: *assertion mode*, *constraints*, and *query mode*.

### 5.4.1 Assertion Mode

This provides an interactive environment in which one can define objects and their types. There are nine basic types corresponding to nine primitives: ∼IND (individuals), ∼TIM (times), ∼LOC (places), ∼REL (relations), ∼POL (polarities), ∼INF (infons), ∼PAR (parameters), ∼SIT (situations), and ∼TYP (types). For instance, if $l$ is a place, then $l$ is of type ∼LOC, and the infon ≪*of-type*, $l$, ∼LOC, 1≫ is a fact in $w$. Note that the type of all types is ∼TYP. For example, the infons ≪*of-type*, ∼LOC, ∼TYP, 1≫ and ≪*of-type*, ∼TYP, ∼TYP, 1≫ are facts in $w$. The syntax of the assertion mode (cf. [26]) is similar to that in [12]. The architecture of Assertion Mode is shown in Figure 4.

Suppose `bob` is an individual, and `sit1` is a situation. Then, these objects can be declared as:

```
I> bob:∼IND
```

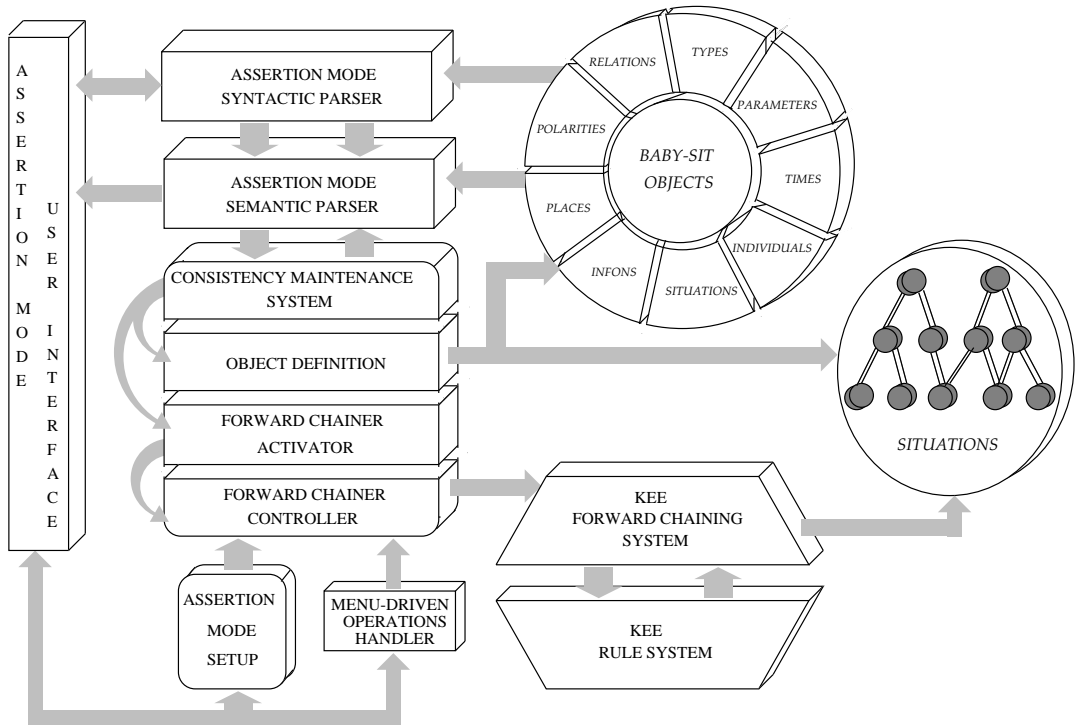

Figure 4: The architecture of Assertion Mode.

    `I> sit1:~SIT`

The definition of relations includes the *appropriateness conditions* for their argument roles. Appropriateness conditions define the domains to which arguments of a relation belong. Each argument can be declared to be from one or more of the primitives above. If we want *sees* to have two arguments, the former being of type individual and the latter being of type situation, we write:

    `I> <sees | ~IND, ~SIT> [1]`

The number in square brackets indicates the minimum number of arguments that can be used with the relation. Hence, ≪sees,bob,1≫, for example, is considered to be a valid (i.e., *unsaturated*) infon in the system and it is equivalent to ≪sees,bob,-,1≫ where "-" is a *null object*.

In order for the parameters to be anchored to objects of the appropriate type, parameters must be declared to be from only one of the primitives. It is also possible to put restrictions on a parameter in the environment. Suppose we want to have a parameter `E` denoting any individual that sees `sit1`. This can be done by asserting:

    `I> E = IND1 ^ ≪sees,IND1,sit1,1≫`

`IND1` is a default system parameter of type ∼IND. `E` is considered as an object of type ∼PAR such that if it is anchored to an object, say `obj1`, then `obj1` must be of type ∼IND and $w$ must support ≪sees,obj1,sit1,1≫.

93

Anchoring situations are those that support infons having *anchor* as their relations. *anchor* has two argument roles: one for a parameter and another for an object which serves as an anchor for the parameter, with the minimality condition 2. For example, if it is the case for Bob that $w \models \ll$sees, bob, sit1,1$\gg$, then an anchoring situation, say `anchor1`, can supply an anchoring which anchors parameter E to `bob`:

   `anchor1` $\models$ $\ll$anchor,E,bob,1$\gg$.

Given an anchoring situation, the anchoring defined by this situation can be applied on the propositions asserted to the system in Assertion Mode. Anchoring is performed by replacing each occurrence of an anchored parameter by its anchor defined in the anchoring situation. For example, giving `anchor1` as the anchoring situation and asserting the proposition

   `sit1` $\models$ $\ll$man,E,1$\gg$

results in the proposition `sit1` $\models$ $\ll$man,bob,1$\gg$ holding.

*Parametric* types are also allowed in BABY-SIT. They are are of the form $[P \mid s \models I]$ where $P$ is a parameter, $s$ is a situation (i.e., a *grounding* situation), and $I$ is a set of infons. The type of all situations that Bob sees can be defined in BABY-SIT as follows:

   I> ~SITALL = [SIT1 | w $\models$ $\ll$sees,bob,SIT1,1$\gg$]

Hence, ~SITALL is seen as an object of type ~TYP in BABY-SIT and can be used as a type specifier for declaration of new objects in the environment. An object of type ~SITALL, say `obj2`, is an object of basic type ~SIT such that $w$ supports the infon $\ll$sees,bob,obj2,1$\gg$.

Naming infons enables one to easily refer to them in expressions. For instance, the infon $\ll$sees,bob,sit1,1$\gg$ can be named `infon1`:

   I> infon1 = $\ll$sees,bob,sit1,1$\gg$

In addition to defining situations, one can create hierarchies among situations. For example, the following sequence of assertions creates a situation `sit2`, defines it as a subsituation of situation `sit1`, and at the same time adds the infon $\ll$blind,bob,0$\gg$ into it:

   I> sit2: ~SIT
   I> sit2 $\models$ {$\ll$make-part-of,sit2,sit1,1$\gg$, $\ll$blind,bob,0$\gg$}

This will force `sit1` to support all the infons supported by `sit2`. As a result, it will be the case that `sit1` $\models$ $\ll$part-of,sit2,sit1,1$\gg$.

Similar operations can be done via the *situation browser* as well. The situation browser enables one to create situations, browse them graphically, add or delete infons, and establish hierarchies among situations. Since all situations are required to cohere in BABY-SIT, assertion of propositions that will yield incoherent situations are refused by the system both for assertions of propositions in Assertion Mode and for the ones asserted during chaining.

Variables in BABY-SIT are solely used in constraints and query expressions, and have scope only within the constraint or the query expression they appear. A variable can match any object appropriate for the place or the argument role it appears in. For example, given the relation above, variables ?S and ?X in the proposition ?S$\models$≪sees,?X,sit1,1≫ can only match objects of type ∼SIT and ∼IND, respectively.

### 5.4.2 Constraints

A BABY-SIT constraint is of the form:

> antecedent$_1$, ..., antecedent$_n$ {<=, =>, <=>}
> consequent$_1$, ..., consequent$_m$.

Each antecedent$_i$, $1 \leq i \leq n$, and each consequent$_j$, $1 \leq j \leq m$, is of the form sit $\{\models, \not\models\}$ ≪rel, arg$_1$, ..., arg$_l$, pol≫ such that rel and each arg$_k$, $1 \leq k \leq l$, can either be an object of the appropriate type or a variable.

Each constraint has an identifier associated with it and must belong to a group of constraints (i.e., a *perspectivity set*). For example, the following is a backward-chaining constraint named HUMAN-BEINGS-012 under the constraint group SPECIES-PERSPECTIVE:

> SPECIES-PERSPECTIVE:
>   HUMAN-BEINGS-012:
>     ?S $\models$ ≪human,?X,1≫ <= ?S $\models$ ≪man,?X,1≫

where ?S and ?X are variables. ?S can only be assigned an object of type ∼SIT while ?X can have values of some type appropriate for the argument roles of the human and man relations. This constraint can apply in any situation. Constraints can also be situated. For example, HUMAN-BEINGS-012 can be rewritten to apply only in situation sit1:

> sit1 $\models$ ≪human,?X,1≫ <= sit1 $\models$ ≪man,?X,1≫.

Conditional constraints of BABY-SIT come with a set of *background conditions* which must be satisfied for the constraint to apply. For example, to state that blocks fall if not supported, one can write:

> NATURAL-LAW-PERSPECTIVE:
>   FALLING-BLOCK:
>     ?S1 $\models$ ≪block,?X,1≫,
>     ?S1 $\models$ ≪supported,?X,0≫ => ?S2 $\models$ ≪falls,?X,1≫
>   UNDER-CONDITIONS:
>     w: ≪exists,gravity,1≫.

Background conditions are assumptions which are required to hold for constraints to be eligible for activation. FALLING-BLOCK can become a candidate for activation only if it is the case that $w \not\models$ ≪exists,gravity,0≫, i.e., if the absence of gravity is not known in the background situation.

The forward-chaining mechanism of BABY-SIT is initiated either when the user tells the system to do so or by assertion of a new object into the system. A candidate forward-chaining constraint is activated whenever its antecedent is satisfied. All the consequences are asserted if they do not yield a contradiction in the situation into which they are asserted. New assertions may in turn activate other forward-chaining constraints. Candidate backward-chaining constraints are activated either when a query is entered explicitly or is issued by the forward-chaining mechanism. Antecedent parts of any constraint are, by default, proved by the backward-chaining constraints in the perspectivity set of that constraint. However, they may also be proved with respect to the backward-chaining constraints in a given *antecedent perspectivity set*.

### 5.4.3 Query Mode

*Query mode* enables one to issue queries about situations. There are several possible actions which can be controlled by the user:

- Providing a perspectivity set to make the query mechanism prove the query with respect to the backward chaining constraints in this set.

- Providing an antecedent perspectivity set to make the query mechanism prove the antecedents of the backward chaining constraints with respect to the backward chaining constraints in this set.

- Searching for solutions by using a given group of constraints.

- Replacing each parameter in the query expression by the corresponding individual if there is a possible anchor, either partial or full, for that parameter provided by the given anchoring situation.

- Returning solutions. (Their number is determined by the user.)

- Displaying a solution with its parameters replaced by the individuals to which they are anchored by the given anchoring situation.

- For each solution, displaying infons anchoring any parameter in the solution to an individual in the given anchoring situation.

- Displaying a trace of anchoring of parameters in each solution.

The computation upon issuing a query is done either by direct querying through situations or by the application of backward-chaining constraints. A situation, $s$, *supports* an infon if the infon is either explicitly asserted to hold in $s$, or it is supported by a situation $s'$ which is part of $s$, or it can be proven to hold by application of backward-chaining constraints. Assume the following:

$\quad$ `sit1` $\models$ {≪sees,E,sit2,1≫, ≪part-of,sit2,sit1,1≫}

$\quad$ `sit2` $\models$ ≪time-of,sit2,t2,1≫

$\quad$ `w` $\models$ ≪sees,bob,sit1,1≫

Given anchoring situation `anchor1`, a query and the system's response to it are as follows:

```
Q> ?S ⊨ {≪sees,E,?Y,1≫, ≪time-of,sit2,?Z,1≫},
   ?S ⊭ ≪blind,bob,1≫
   Solution 1:
   sit1 ⊨ {≪sees,bob,sit2,1≫, ≪time-of,sit2,t2,1≫},
    sit1 ⊭ ≪blind,bob,1≫
   with the anchoring:
   anchor1 ⊨ ≪anchor,E,bob,1≫.
```

## 5.5 Critique of PROSIT and ASTL

A tableau comparison of PROSIT, ASTL, and BABY-SIT is given in Table 1.

### 5.5.1 Types

At the heart of situation theory lies a scheme of individuation. Situations, relations, individuals, temporal locations, and spatial locations are the basic uniformities. The need for a mathematical representation of these uniformities resulted in what is known as *types*. Types are higher-order uniformities which cut across basic uniformities. The ontology of situation theory has been extended further to include other uniformities such as infons, polarities, etc. In this respect, PROSIT and ASTL do not allow their objects to be of some type. Only situations can be declared to have a situation type. Other objects in the system are left untyped. This approach has particular consequences on the conception of relations and parameters which are explained in the sequel.

### 5.5.2 Parameters

The development of types necessitates devices, such as parameters, for making reference to arbitrary objects of a given type. In ASTL, there is no special treatment of parameters which are just atomic objects in the model. Declaring situations to be of some type allows abstraction over situations to some degree. But, the actual means of abstraction over objects in situation theory, viz. parameters, does not carry much significance in ASTL. Parameters are only used in identifying situation types. Since there is no notion of types other than situation types in ASTL, a parameter can hold the place of any object. PROSIT treats parameters in a way similar to its variables, except they can be equated to any constant or parameter. PROSIT has no mechanism to define types. It cannot define a situation-type explicitly. On the other hand PROSIT can query a certain type of situation and put constraints between situation-types.

| Constraint Type | PROSIT | ASTL | BABY-SIT |
|---|---|---|---|
| Nomic | √ | √ | √ |
| Necessary | √ | √ | √ |
| Conditional | – | – | √ |
| Situated | √ | – | √ |
| Global | – | √ | √ |

| Constraint Class | PROSIT | ASTL | BABY-SIT |
|---|---|---|---|
| Situation constraint | – | √ | √ |
| Infon constraint | √ | √ | √ |
| Argument constraint | – | – | √ |

| Computational Feature | PROSIT | ASTL | BABY-SIT |
|---|---|---|---|
| Unification | √ | √ | √ |
| Type-theoretic | – | – | √ |
| Coherence | – | – | √ |
| Forward-chaining | √ | – | √ |
| Backward-chaining | √ | √ | √ |
| Bidirectional-chaining | √ | – | √ |

| Miscellaneous Features | PROSIT | ASTL | BABY-SIT |
|---|---|---|---|
| Circularity | √ | √ | √ |
| Partiality | √ | √ | √ |
| Parameters | ? | ? | √ |
| Type Abstraction | ? | ? | √ |
| Parameter restriction | – | ? | √ |
| Anchoring | ? | ? | √ |
| Information nesting | √ | √ | √ |
| Unsaturated infons | ? | – | √ |
| Nonmonotonic reasoning | – | – | √ |
| Set operations | √ | – | – |

Legend: √: exists, –: doesn't exist,
?: partially/conceptually exists.

Table 1: Tableau comparison of existing approaches.

It is useful to have parameters that range over various classes rather than to work with parameters ranging over all objects. Such particularized parameters can be obtained by parameter restriction. On the other hand, in situation theory, parameters are used to achieve abstraction at the level of almost all object types, i.e., situations, individuals, temporal locations, etc. by using type abstraction.

In PROSIT some of these are hard to achieve. First of all, there is no typing in PROSIT. A variable can match any parameter or constant without due regard to types. Obtaining restricted parameters and type abstraction is not possible since there is no built-in mechanism in the system. But one can pose queries on restricted parameters. For example, all men kicking footballs can be queried using the following expression:

```
(AND (kicking *a *b) (man *a) (football *b)).
```

Although none of the variables are restricted, the expression queries a restricted class of individuals.

In ASTL, abstraction is only at the level of situations. There is no direct equivalent of properties in ASTL. Consider the abstraction for an individual having the property of being happy in some situation $s$:

$$[X \mid s \models \ll happy, X, 1 \gg].$$

In ASTL, Black achieves this by allowing situation types with parametric infons. But this is not an appropriate way to use abstractions since one cannot abstract over other objects such as individuals, temporal locations, etc. (cf. *object type-abstraction* and *situation type-abstraction* in [12]).

### 5.5.3 Parameter Anchoring

Parameters are place holders for indeterminate objects in situation theory and yield a form of abstraction over objects. The ties of these abstractions with the real world occur via a kind of assignment function called *anchor*. This function changes from one cognitive agent to another, and from one perspective to another of a single cognitive agent. Information content of an abstract object increases when its parameters are anchored to objects in the real world by an anchor. An anchor maps a parameter to a unique, appropriate object in the world. Technically speaking, a parameter must be anchored to an object of the same type since the parameter is a filler for an object having specific properties. The issues of anchoring to a unique object and anchoring to an object of the same type introduce technical difficulties in building a computational system.

Some treatment of parameters is given in PROSIT with respect to *anchoring*. Given a parameter denoting an object of some type (individual, situation, etc.), an anchor is a function which assigns an object of the same type to the parameter [12, pp. 52–63]. Hence, parameters work by placing restrictions on anchors. But, there is no appropriate anchoring mechanism in PROSIT since its parameters are untyped.

In the case of ASTL, there are several points worth mentioning. Black proposes to consider anchors as situations (*anchoring situations*) having infons of the form ≪*anchor-to*, *label*, *term*≫ and other related infons. Second, the current version of ASTL must be modified to use anchoring situations. This cannot be controlled by the user. The main reason is that whenever an anchoring occurs, the system must check whether the first argument of the relation *anchor-to* is a label and the second one is a term. Moreover, the system must assure that the parameter is anchored to only one object in that anchoring situation. Finally, type checking for both of the arguments is required. The crux of all these problems lies in ASTL's not having type-theoretic objects and not employing parameters as they are intended in situation theory.

### 5.5.4 Infons

There are three characteristics of an infon in the existing systems which should be evaluated from the standpoint of situation theory: *argument places*, *minimality conditions*, and *argument roles*.

Each relation should have a limited number of argument places. Consider the relation *walks*. A reasonable assumption is that this relation has four argument places: a walking agent, a direction/destination, the location of walking, and the time of walking.

To have a formally well-defined infon, there must be a lower bound as to the number of argument places to be filled in an $n$-place relation. For example, at least one argument place of the relation *walks* is to be filled, namely the walking agent. Otherwise, the infon ≪*walks*≫ would have zero information content. Minimality conditions are, then, necessary for a relation to provide an item of information. All argument places of a relation in ASTL are required to be filled, and consequently all infons are *saturated*. As for the infons in PROSIT, there is no restriction as to the number of argument roles of a relation to be filled.

Any object appearing as an argument of a relation must be appropriate for the argument role imposed by that argument place. Hence, *appropriateness* conditions must be defined for each possible argument place of a relation. This is generally done by forming a set of infons for an argument place which are supposed to be supported by the *world situation* for a given object. At the primary level, each argument role requires the appropriate object to be of some basic type. That is, each argument role is associated with a certain *type*, the type of the object that may legitimately fill that argument role. In a technical sense, appropriate conditions for an argument role are *complex* types having possibly the world situation as their *grounding* situation.

PROSIT and ASTL do not allow definition of appropriateness conditions for arguments of relations, mainly because objects are not typed in these systems. However, one can define restrictions on the parameters of infons by using restricted infons in PROSIT. The relation *walks*, for example, might require its walking agent role to be filled by an animate object. Such a restriction can be defined only by using constraints in PROSIT and ASTL. However, this requires writing the restriction each time a new constraint about *walks* is to be added. Having appropriateness conditions as a built-in feature would be better.

### 5.5.5 Hierarchy of Situations

Being in a larger situation gives one the ability of having information about its subsituations. Although there is no mention of hierarchy in situation theory, the *part-of* relation can be used to build such a structure (i.e., *information nesting*) among abstract situations. ASTL does not have a mechanism to relate two situations so that one will directly support all the facts that the other does. While this might be achieved via constraints in ASTL, there is no built-in structure between situations.

PROSIT has a tree structure among situations established by the use of owner and subchunk relations. In fact, this hierarchy of PROSIT turns out to be useful in problems regarding knowledge and belief.

The other two PROSIT relations (subtype and subsituation) should be examined carefully. At first glance, it seems that there is a similarity between these relations and the concept of inheritance in object-oriented programming. However, in PROSIT the supersituation inherits all the infons from the subsituation, whereas in object-oriented programming it is the subclass that inherits the properties and methods from the superclass.

Another question may come as to where one can use these relations. The example given in the PROSIT manual uses these relations to classify the airplanes of type DC (DC–9, DC–10, and so on). But from the situation-theoretic point of view, it is not correct to consider airplanes of type DC as a situation. An agent does not individuate DC type of airplanes as a situation and DC–9s as a subsituation of that situation. These can only be considered as a class and its subclass. This example is surely well suited to object-oriented programming, but not to situation theory.

### 5.5.6 Coherence of Situations

ASTL does not provide a mechanism, such as truth maintenance, to preserve coherence within situations. This is left to the user's control and can be achieved by specifying some special constraints in the ASTL descriptions. A constraint of the form

```
*S: [S | S |= ⟨actual,S,0⟩] -> *S: [S | S |= ⟨*R,*A,1⟩],
                              *S: [S | S |= ⟨*R,*A,0⟩]
```

is given by Black as an example. However, this is not a solution to the problem of having *incoherent situations*. Moreover, this approach may be quite expensive for the user since maintaining coherence is a complicated task and when left to the user, a large number of constraints must be written. What is worse is that consequences of allowing incoherent situations and reasoning over them may be drastic, e.g., it may lead to unintended models during computation. It seems that coherence, as a built-in notion, can hardly be embedded in an extension of the existing version of ASTL since it is not a syntactical matter and requires meta level control over the whole system.

Similar to ASTL, PROSIT cares little about coherence within situations. This is left to the user's control.

### 5.5.7 Constraints

PROSIT supports the concept of constraints, but handles them in a different fashion. These come in three flavors in PROSIT: forward-chaining constraints, backward-chaining constraints, and forward- and backward-chaining constraints (bidirectional-chaining constraints). In fact, both methods (forward or backward) result in the same answers to queries. However, forward-chaining incurs a high cost at assertion-time, and backward-chaining incurs a high cost at query-time.

ASTL constraints are all in the form of backward-chaining constraints. The user can only issue queries. However, an intelligent agent has the ability to not only acquire information about situations and obtain new information about them by being attuned to assorted constraints, but also act accordingly to alter its environment. Thus, having forward-chaining constraints as well would be better. In this way, new situations would be created, new infons would be inserted into situations, and consequences of new infons would be observed.

PROSIT's constraints are situated infon constraints, i.e., they are about local facts within a situation rather than about situation-types. Still, it is possible to simulate constraints that are not local to one situation (but are global). This can be achieved by introducing a situation which is global to all other situations and then asserting the constraint in this global situation. Because all other situations will be in this global situation, any constraint that is asserted here will apply to all situations. For example,

```
(!= (resp topsit
  (<= (!= *Sit1 (touching *X *Y))
      (!= *Sit1 (kissing *X *Y)))))
```

states that if, in situation `topsit`, there is a situation that supports a fact with the relation *kissing*, then that situation also supports a fact with the relation *touching* on the same arguments.

Situated constraints offer an elegant solution to the treatment of conditional constraints which apply in situations that obey some condition. For example, when Alice throws a basketball, she knows it will come down—a constraint to which she is attuned, but which would fail if she tried to play basketball in a space shuttle. This is actually achieved in PROSIT since information is specified in the constraint itself. Situating a constraint means that it may only apply to appropriate situations. This is a good strategy to achieve *background conditions*. However, it might be required that conditions set not only within the same situation, but also between various types of situations. Because constraints have to be situated in PROSIT, not all situations of the appropriate type will have a constraint to apply.

Although one can define constraints between situations in ASTL, the notion of background conditions for constraints is not available. This means that *conditional constraints* are not available. However, this can be achieved by writing a set of conditions which must be satisfied for the constraint to qualify as an applicable one. These conditions will obviously be placed on the consequent part of each ASTL constraint since all ASTL constraints are used for backward-chaining.

Black identifies three classes of constraints in [8]:

- Situation constraints: Constraints between situation types.
- Infon constraints: Constraints between infons (of a situation).
- Argument constraints: Constraints on argument roles (of an infon).

Only PROSIT cannot model situation constraints since it does not have situation types. Defining infon constraints is possible in all systems. However, argument constraints are a built-in feature only in BABY-SIT since they directly correspond to having appropriateness conditions for argument roles of relations in infons.

### 5.5.8 Nonmonotonicity

A typical user studying situation theory will not only want to investigate if an infon is supported by a situation, but also want to see if an infon is not supported by that situation. In other words, he would like to know if a situation is not of a certain type and then use this knowledge. This calls for negation in both query statements and constraints. A straightforward way to do this is by having the appropriate syntax and semantics for the negation of *supports* relation, i.e., by letting "$\not\models$" be used in these statements. Consider the BABY-SIT constraint:

```
?S |= ≪paid-little,?W,?S,1≫,
?S |≠ ≪has-other-income,?W,?S,1≫ <= ?S |= ≪poor,?W,1≫
```

which expresses the rule of reasoning "a worker is poor if he is paid little, under the assumption that he has no other income."

Having such a construct in the constraint mechanism, and hence in the query mechanism, allows nonmonotonic reasoning. Unfortunately, neither PROSIT nor ASTL have an equivalent construct.

### 5.5.9 Some Formal Properties

Black shows that ASTL is sound, but he leaves its completeness formally unproved. Similar arguments are valid for PROSIT as well. Although it has not been proved explicitly, PROSIT can be said to be a sound system. BABY-SIT is, on the other hand, a sound and complete system. BABY-SIT is based on the constructs and the inference mechanism of KEE. The situation structures are developed upon the KEE's *world system* based on Morris and Nado's work [21]. The inference mechanism of the BABY-SIT is the same as that of KEE, except for the chaining control mechanism, and BABY-SIT constraints form a subset of the *action rules* that can be defined in KEE.

### 5.5.10 Domains of Application

The main group of problems that PROSIT can handle is that of individual knowledge and belief in multi-agent systems, and common knowledge (mutual information). There are three main properties that enable PROSIT to simulate human-like reasoning. The first one is situated programming, i.e., infons and constraints are local to situations. The second is PROSIT's situation tree structure, which one can use to represent nested knowledge/belief. The third is the use of inconsistencies to generate new information. Self-referential expressions and situations as arguments of infons are two powerful features. These features can efficiently be used in representing knowledge and belief. The owner relation and the super-chunk relation are useful in modeling epistemic puzzles [14].

ASTL has been developed with natural language processing in mind. Still, it is possible to use it as a general knowledge representation language. The advantages of employing declarative or procedural approaches in knowledge-based systems are still being debated. Both have been justified from perspectives of cognitive science and philosophy. For the time being, the declarative approach fits best for a situation-theoretic computational language, but one can also benefit from procedural knowledge. PROSIT is a candidate for a unified framework since it is possible to use Lisp statements as part of the language.

BABY-SIT is being developed as a general-purpose programming environment, specifically adopting the ontological features of situation theory and putting them into the comfortable reach of the user. Its interactive nature and facilities to organize and keep track of information qualifies it as a general knowledge representation system. The flexibility of situation semantics as a powerful linguistic account to handle various linguistic phenomena has led to the initiation of a preliminary study towards employing BABY-SIT in the resolution of pronominal anaphora [30, 31].

### 5.5.11 User Interfaces

PROSIT and ASTL provide simple user interfaces. The user writes definitions into a file which can be loaded in a Common Lisp environment. Other than querying what situations support, the user has the opportunity to view some system features. ASTL is not an interactive language in the sense that a static definition is input to the system and the user can only observe what can be inferred from these definitions. Moreover, one cannot assert propositions to the system; new propositions must first be added to the static description and then the system must be reloaded. This prevents the user from directly seeing the consequences of his propositions.

# 6 Concluding Remarks

Serious thinking about the computational aspects of situation theory is just starting. There have been some proposals [8, 9, 16, 26] in this direction, with varying degrees of divergence from the ontology of situation theory. ASTL [8] and PROSIT [9] mainly offer a Prolog- or Lisp-like programming language while BABY-SIT [26, 32] provides a programming environment incorporating situation-theoretic constructs.

We believe that computational aspects of situation theory call for deeper investigation. Although the current attempts are in their infancy, they already have some applicability in artificial intelligence and natural language processing. However, their use should be further demonstrated to show why situation theory provides a challenging arena for studying various phenomena in these fields.

# Acknowledgments

We gratefully acknowledge Erhard W. Hinrichs and Tsuneko Nakazawa, the organizers of the Workshop on "Grammar Formalisms for Natural Language Processing," for providing the funding for the presentation of this paper in the *Sixth European Summer School on Logic, Language and Information*, Copenhagen, Denmark, August 1994.

We are also indebted to John Griffith for his strong initiative to publish this technical report.